\documentclass[11pt,a4paper]{article}
\usepackage[hyperref]{acl2017}
\usepackage{times}
\usepackage{latexsym}
\usepackage{graphicx}

\usepackage{url}

\newcommand{\mltpl}[2]{$#1#2$}
\newcommand{\edit}[1]{{\color{black} #1}}

\aclfinalcopy 

\title{Best--Worst Scaling More Reliable than Rating Scales:\\ {\rm A Case Study on Sentiment Intensity Annotation}}

\author{Svetlana Kiritchenko \and Saif M. Mohammad\\
	    National Research Council Canada\\
	    {\tt \small \{svetlana.kiritchenko,saif.mohammad\}@nrc-cnrc.gc.ca}
}

\date{}

\begin{document}
\maketitle
\begin{abstract}
  Rating scales are a widely used method for data annotation; 
  however, they present several
  challenges, such as difficulty in maintaining inter- and intra-annotator consistency. 
  Best--worst scaling (BWS) is an alternative method of annotation
  that is claimed to produce high-quality annotations while keeping the required number of annotations similar to that of rating scales.  
  However, the veracity of this claim has never been systematically established.  
  Here for the first time, we set up an experiment
  that directly compares the rating scale method with BWS. We show that with the same total number of annotations, BWS produces significantly more reliable results than the rating scale.
\end{abstract}

\section{Introduction}

When manually annotating data with quantitative or qualitative information, researchers in many disciplines, including social sciences and computational linguistics, often rely on {\it rating scales (RS)}. 
A rating scale provides the annotator with a choice of categorical or numerical values that represent the measurable characteristic of the rated data. 
For example, when annotating a word for sentiment, the annotator can be asked to choose among integer values from 1 to 9, with 1 representing the strongest negative sentiment, and 9 representing the strongest positive sentiment \cite{bradley1999affective,warriner2013norms}. 
Another example is the Likert scale, which measures responses on a symmetric agree--disagree scale, from `strongly disagree' to `strongly agree' \cite{likert1932technique}. 
The annotations for an item from multiple respondents are usually averaged to obtain a real-valued score for that item.
Thus, for an $N$-item set, if each item is to be annotated by five respondents, then the number of annotations required is \mltpl{5}{N}.

While frequently used in many disciplines, the rating scale method has a number of limitations \cite{presser2004questions,baumgartner2001response}. These include:
\vspace*{-2mm}
\begin{itemize}
\item {\it Inconsistencies in annotations by different annotators}: one annotator might assign a score of 7 to the word \textit{good} on a 1-to-9 sentiment scale, while another annotator can assign a score of 8 to the same word.
\vspace*{-3mm}
\item {\it Inconsistencies in annotations by the same annotator}: an annotator might assign different scores to the same item when the annotations are spread over time.
\vspace*{-3mm}
\item {\it Scale region bias}: annotators often have a bias towards a part of the scale, for example, 
preference for the middle of the scale.
\vspace*{-3mm}
\item {\it Fixed granularity:} 
in some cases, annotators might feel too restricted with a given rating scale and may want to place an item in-between the two points on the scale. On the other hand, a fine-grained scale may overwhelm the respondents and lead to even more inconsistencies in annotation.
\end{itemize}
\vspace*{-2mm}

{\it Paired Comparisons} \cite{thurstone1927law,david1963method} is a comparative annotation method, 
where respondents are presented with pairs of items and asked which item has more of the property of interest (for example, which is more positive). The annotations can then be converted into a ranking of items by the property of interest, and one can even obtain real-valued scores indicating the degree to which an item is associated with the property of interest.
The paired comparison method does not suffer from the problems discussed above for the rating scale, but it requires a large number of annotations---order $N^2$, where $N$ is the number of items to be annotated. 

{\it Best--Worst Scaling (BWS)} is a less-known, and more recently introduced, variant of comparative annotation. It was developed by \citet{Louviere_1991}, building on some groundbreaking research in the 1960's in mathematical psychology and psychophysics by Anthony A. J. Marley and Duncan Luce. 
Annotators are presented with $n$ items at a time (an $n$-tuple, where $n > 1$, and typically $n = 4$). 
They are asked which item is the {\it best} (highest in terms of the property of interest) and which is the {\it worst} (lowest in terms of the property of interest). 
When working on 4-tuples, best--worst annotations are particularly efficient
because by answering these two questions, the results for five out of six item--item pair-wise comparisons become known. 
All items to be rated are organized in a set of $m$ 4-tuples ($m \geq N$, where $N$ is the number of items) so that each item is evaluated several times in diverse 4-tuples. 
Once the $m$ 4-tuples are annotated, one can compute real-valued scores for each of the items 
using a simple counting procedure \cite{Orme_2009}.  
The scores can be used to rank items by the property of interest.

BWS is claimed to 
produce high-quality annotations while still keeping the number of 
annotations small (\mltpl{1.5}{N}--\mltpl{2}{N} tuples need to be annotated) \cite{Louviere2015,maxdiff-naacl2016}. However, the veracity of this claim has never been systematically established.  
In this paper, we pit the widely used rating scale squarely against BWS in a quantitative experiment to determine which method provides more reliable results. We produce real-valued sentiment intensity ratings for 3,207 English terms (words and phrases) using both methods by aggregating responses from several independent annotators. 
We show that BWS 
ranks terms more reliably, that is, when
 comparing the term rankings obtained from two groups of annotators for the same set of terms, the correlation between the two sets of ranks produced by BWS is significantly higher than the correlation for the two sets obtained with RS. 
The difference in reliability is more marked when about 5$N$ (or less) total annotations are obtained, 
which is the case in many NLP annotation projects \cite{strapparava-mihalcea:2007:SemEval-2007,Socher2013,MohammadT13}. 
Furthermore, 
the reliability obtained by rating scale when using 
ten annotations per term 
is matched by BWS with only \mltpl{3}{N} total annotations (two annotations for each of the \mltpl{1.5}{N} 4-tuples).

The sparse prior work in natural language annotations that uses BWS 
involves the creation of datasets for relational similarity \cite{jurgens-EtAl:2012:STARSEM-SEMEVAL}, word-sense disambiguation \cite{Jurgens2013EmbracingAA}, and word--sentiment intensity \cite{maxdiff-naacl2016}.
However, none of these works has systematically compared BWS with the rating scale method. 
We hope that our findings will encourage the use of BWS more widely to obtain high-quality NLP annotations. 
All data from our experiments as well as scripts to generate BWS tuples, to generate item scores from BWS annotations, and for assessing reliability of the annotations are made freely available.\footnote{www.saifmohammad.com/WebPages/BestWorst.html}

\section{\scalebox{.98}{Complexities of Comparative Evaluation}}
Both rating scale and BWS are less than perfect ways to capture the true word--sentiment intensities in the minds of native speakers of a language. Since the ``true" intensities are not known, determining which approach is better is non-trivial.\footnote{Existing sentiment lexicons are a result of one
or the other method and so cannot be treated as the truth.}

A useful measure of quality is 
reproducibility---if repeated independent manual annotations from multiple respondents result in similar sentiment scores, then one can be confident that the scores capture the true sentiment intensities. Thus, we set up an experiment that compares BWS and RS in terms of how similar the results are on repeated independent annotations.

It is expected that reproducibility improves with the number of annotations for both methods. (Estimating a value often stabilizes as the sample size is increased.)   
However, in rating scale annotation, each item is annotated individually whereas in BWS, groups of four items (4-tuples) are annotated together (and each item is present in multiple different 4-tuples).
To make the reproducibility evaluation fair, we ensure that the term scores are inferred from the same total number of annotations for both methods. 
For an $N$-item set, let $k_{rs}$ be the number of times each item is annotated via a rating scale. Then the total number of rating scale annotations is \mltpl{k_{rs}}{N}. 
For BWS, let the same $N$-item set be converted into $m$ 4-tuples that are each annotated $k_{bws}$ times. Then the total number of BWS annotations is \mltpl{k_{bws}}{m}. 
In our experiments, we compare results across BWS and rating scale at points when \mltpl{k_{rs}}{N} $=$ \mltpl{k_{bws}}{m}. 

The cognitive complexity involved in answering a BWS question is different from that in a rating scale question. On the one hand, for BWS, the respondent has to consider four items at a time simultaneously. On the other hand, even though a rating scale question explicitly involves only one item, the respondent must choose a score that places it appropriately with respect to other items.\footnote{A somewhat straightforward example is that {\it good} cannot be given a sentiment score less than what was given to {\it okay}, and it cannot be given a score greater than that given to {\it great}. Often, more complex comparisons need to be considered.} Quantifying the degree of cognitive load of a BWS annotation vs. a rating scale annotation (especially in a crowdsourcing setting) is particularly challenging, and beyond the scope of this paper. 
Here we explore the extent to which the rating scale method and BWS lead to the same resulting scores when the annotations are repeated, controlling for the total number of annotations.

\section{Annotating for Sentiment}

We annotated 3,207 terms for sentiment intensity (or degree of positive or negative valence) with both the rating scale and best--worst scaling.
The annotations were done by crowdsourcing on CrowdFlower.\footnote{\edit{
The full set of annotations as well as the instructions to annotators for both methods are available at http://www.saifmohammad.com/WebPages/BestWorst.html.}}
\edit{The workers were required to be native English speakers from the USA.}

\subsection{Terms}

The term list includes 1,621 positive and negative single words from Osgood's valence subset of the General Inquirer \cite{Stone66}. 
It also included 1,586 high-frequency short phrases formed by these words in combination with simple negators (e.g., {\it no}, {\it don't}, and {\it never}), modals (e.g., {\it can}, {\it might}, and {\it should}), or degree adverbs (e.g., {\it very} and {\it fairly}). 
More details on the term selection can be found in \cite{SCL-NMA}.

\subsection{Annotating with Rating Scale}

The annotators were asked to rate each term on a 9-point scale, ranging from $-4$ (extremely negative) to $4$ (extremely positive). 
The middle point (0) was marked as `not at all positive or negative'. 
Example words were provided for the two extremes ($-4$ and $4$) and the middle (0) to give the annotators a sense of the whole scale.

Each term was annotated by twenty workers for the total number of annotations to be \mltpl{20}{N} ($N = 3,207$ is the number of terms). 
A small portion (5\%) of terms were internally annotated by the authors. 
If a worker's accuracy on these check questions fell below 70\%, that worker was refused further annotation, and all of their responses were discarded.
The final score for each term was set to the mean of all ratings collected for this term.\footnote{When evaluated as described in Sections~\ref{ann_diff} and \ref{ann_reliability}, median and mode produced worse results than mean.} 
On average, the ratings of a worker correlated well with the mean ratings of the rest of the workers (average Pearson's $r = 0.9$, min $r = 0.8$). Also, the Pearson correlation between the obtained mean ratings and the ratings from similar studies by \citet{warriner2013norms} and by \citet{dodds2011temporal} were 0.94 (on 1,420 common terms) and 0.96 (on 998 common terms), respectively.\footnote{ \edit{\citet{warriner2013norms} list a correlation of 0.95 on 1029 common terms with the lexicon by \citet{bradley1999affective}.}}

\begin{figure}[t]
\centering
\includegraphics[width=3in]{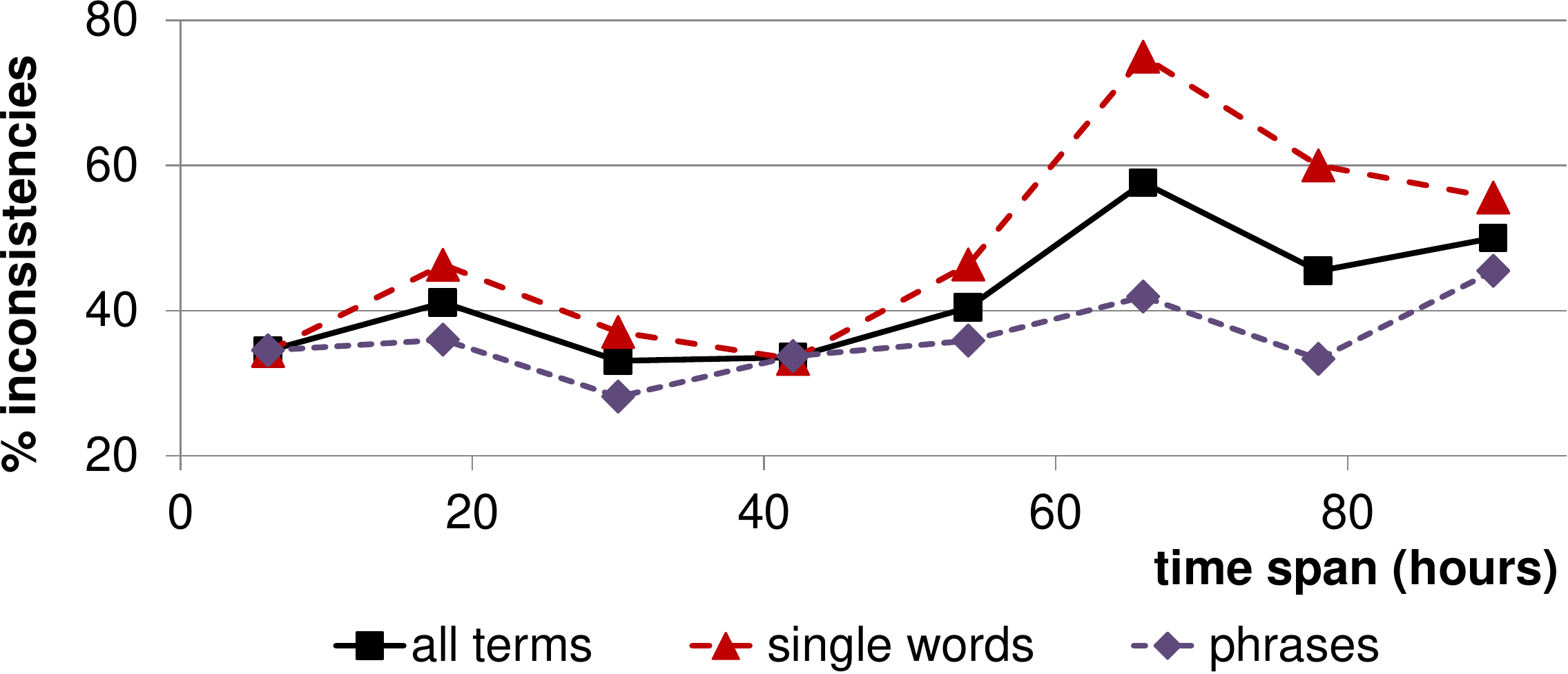}
\caption{The inconsistency rate in repeated annotations by same workers using rating scale.}
\vspace*{-3mm}
\label{fig:RS-incons}
\vspace*{-2mm}
\end{figure}

To determine how consistent individual annotators are over time,
180 terms \edit{(90 single words and 90 phrases)} were presented for annotation twice with intervals ranging from a few minutes to a few days. 
For 37\% of these instances, the annotations for the same term by the same worker were different. 
The average rating difference for these inconsistent annotations was 1.27 (on a scale from $-4$ to $4$).  
Fig.~\ref{fig:RS-incons} shows the inconsistency rate in these repeated annotations as a function of time interval between the two annotations. 
The inconsistency rate is averaged over 12-hour periods. 
One can observe that intra-annotator inconsistency increases with the increase in time span between the annotations. 
Single words tend to be annotated with higher inconsistency than phrases. 
However, when annotated inconsistently, phrases have larger average difference between the scores (1.28 for phrases vs. 1.21 for single words). 
Twelve out of 90 phrases (13\%) have the average difference greater than or equal to 2 points. 
This shows that it is difficult for annotators to remain consistent when using the rating scale.

\subsection{Annotating with Best--Worst Scaling}

The annotators were presented with four terms at a time (a 4-tuple) and asked to select the most positive term and the most negative term. 
The same quality control mechanism   
of assessing a worker's accuracy on internally annotated check questions 
(discussed in the previous section)
was employed here as well. 
\mltpl{2}{N} (where $N = 3,207$) 
distinct 4-tuples were randomly generated in such a manner that each term was seen in eight different 4-tuples, and no term appeared more than once in a tuple.\footnote{The script used to generate the 4-tuples is available at http://www.saifmohammad.com/WebPages/BestWorst.html.} 
Each 4-tuple was annotated by 10 workers. 
Thus, the total number of annotations obtained for BWS was \mltpl{20}{N} (just as in RS). 
We used the partial sets of \mltpl{1}{N}, \mltpl{1.5}{N}, and the full set of \mltpl{2}{N} 4-tuples to investigate the impact of 
the number of unique 4-tuples
on the quality of the final scores. 

We applied the {\it counting procedure} to obtain real-valued term--sentiment scores from the BWS annotations \cite{Orme_2009,flynn2014}: 
the term's score was calculated as 
the percentage of times the term was chosen as most positive minus the percentage of times the term was chosen as most negative. 
The scores range from $-1$ (most negative) to 1 (most positive). 
This simple and efficient procedure has been shown to produce results similar to ones obtained with more sophisticated statistical models, such as multinomial logistic regression \cite{Louviere2015}.

In a separate study, we use the resulting dataset of 3,207 words and phrases annotated with real-valued sentiment intensity scores by BWS, which we call Sentiment Composition Lexicon for Negators, Modals, and Degree Adverbs (SCL-NMA), to analyze the effect of different modifiers on sentiment \cite{SCL-NMA}.

\section{How different are the results obtained by rating scale and BWS?}
\label{ann_diff}

The difference in final outcomes of BWS and RS can be determined in two ways:
by directly comparing term scores or by comparing term ranks. 
To compare scores, we first linearly transform the BWS and rating scale scores to scores in the range 0 to 1.
Table \ref{tab:diff} shows the differences in scores, differences in rank, Spearman rank correlation $\rho$, and Pearson correlation $r$ for 3$N$, 5$N$, and 20$N$ annotations.  
Observe that the differences are markedly larger for commonly used annotation scenarios where only 3$N$ or 5$N$ total annotations are obtained, but even with 20$N$ annotations, the differences across RS and BWS are notable.

\begin{table}[t]
\begin{center}
\small
\begin{tabular}{rrrrrr}
\hline  \# annotations & $\Delta$ score & $\Delta$ rank & $\rho$ & $r$\\ 
\hline
3$N$ &0.11 &397 & 0.85 & 0.85\\
5$N$ &0.10 &363 &0.87 & 0.88\\
20$N$ &0.08  &264 &0.93 & 0.93\\
\hline
\end{tabular}
\end{center}
\vspace*{-2mm}
\caption{\label{tab:diff} Differences in final outcomes of BWS and RS, for different total numbers of annotations.}
\end{table}

\setlength{\tabcolsep}{5pt}
\begin{table}[t]
\begin{center}
\small
\begin{tabular}{lrrr}
\hline \bf Term set & \bf \# terms & \multicolumn{1}{c}{\bf $\rho$} & \multicolumn{1}{c}{\bf $r$} \\ \hline
all terms & 3,207 & .93 & .93\\[2pt]
single words & 1621 & .94 & .95\\[2pt]
all phrases & 1586 & .92 & .91\\
\ \ negated phrases & 444 & .74 & .79\\
\ \ \ \ pos.\@ phrases that have a negator & 83 & -.05 & -.05\\
 \ \ \ \ neg.\@ phrases that have a negator & 326 & .46 & .46\\
\ \ modal phrases & 418 & .75 & .82\\
\ \ \ \ pos.\@ phrases that have a modal & 272 & .44 & .45\\
\ \ \ \ neg.\@ phrases that have a modal & 95 & .57 & .56\\
\ \ adverb phrases & 724 & .91 & .95\\
\hline
\end{tabular}
\end{center}
\vspace*{-2mm}
\caption{\label{corr-modifiers} Correlations between sentiment scores produced by BWS and rating scale.}
\vspace*{-3mm}
\end{table}

\setlength{\tabcolsep}{6pt}

Table~\ref{corr-modifiers} shows Spearman ($\rho$) and Pearson ($r$) correlation between the ranks and scores produced by RS and BWS on the full set of \mltpl{20}{N} annotations.
Notice that the scores agree more on single terms and less so on phrases. 
The correlation is noticeably lower for phrases involving negations and modal verbs. 
Furthermore, the correlation drops dramatically for positive phrases that have a negator (e.g., \textit{not hurt}, \textit{nothing wrong}).\footnote{A term was considered positive (negative) if the scores obtained for the term with rating scale and BWS are both greater than or equal to zero (less than zero). Some terms were rated inconsistently by the two methods; therefore, the number of the positive and negative terms for a category (negated phrases and modal phrases) does not sum to the total number of terms in the category.} 
The annotators also showed greater inconsistencies while scoring these phrases on the rating scale (std.\@ dev.\@ $\sigma = 1.17$ compared to $\sigma = 0.81$ for the full set).
Thus it seems that the outcomes of rating scale and BWS diverge to a greater extent when the complexity of the items to be rated increases.

\section{Annotation Reliability}
\label{ann_reliability}

\begin{figure}[t]
\centering
\includegraphics[width=3in]{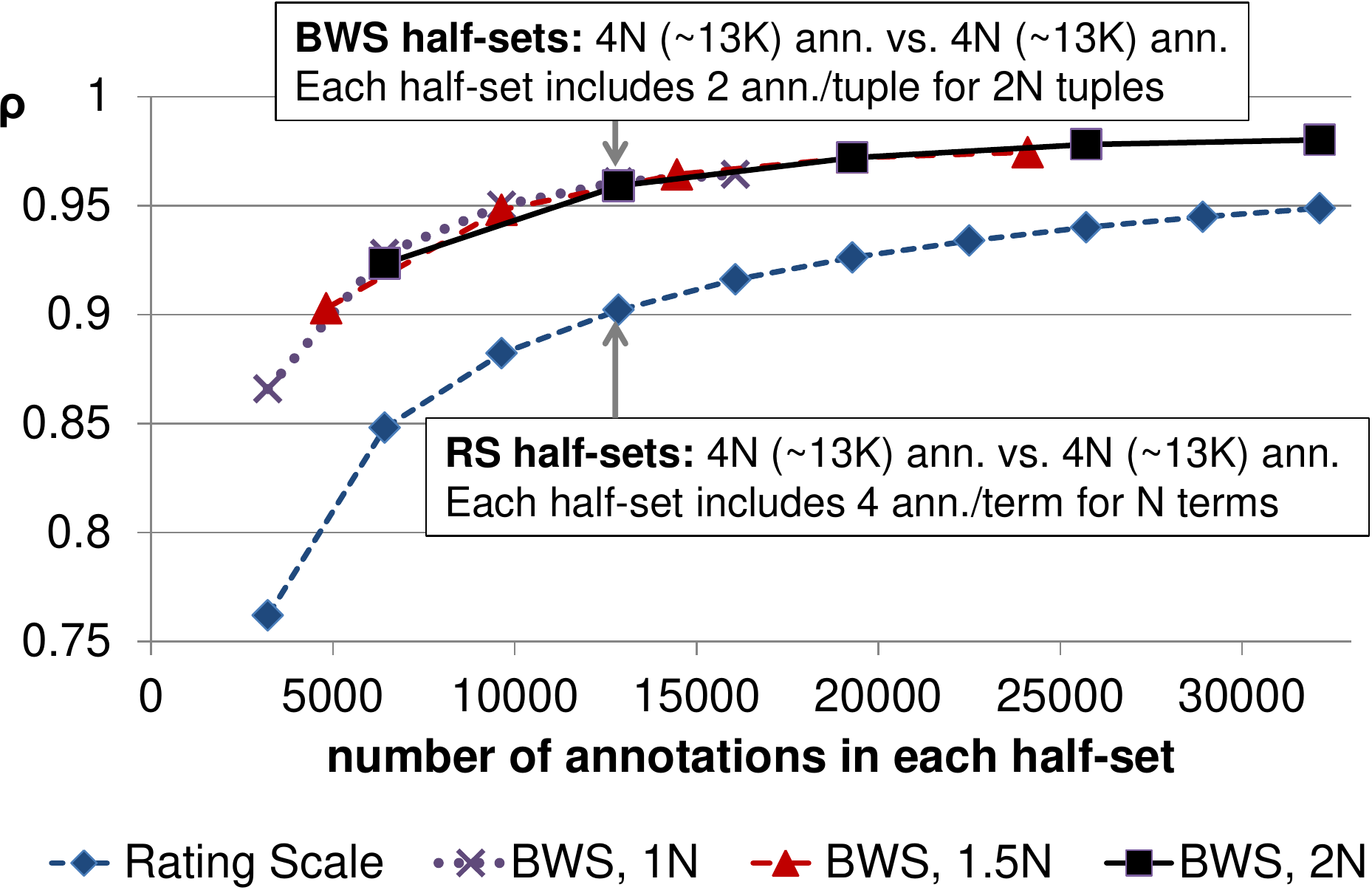}
\vspace*{-2mm}
\caption{SHR for RS and BWS (for $N=3207$).}
\label{fig:Reliability}
\vspace*{-5mm}
\end{figure}

To assess the reliability of annotations produced by a method (BWS or rating scale), we calculate average {\it split-half reliability (SHR)} over 100 trials. SHR is a commonly used approach to determine consistency in psychological studies, that we employ as follows.   
All annotations for a term or a tuple are randomly split into two halves. Two sets of scores are produced independently from the two halves. 
Then the correlation between the two sets of scores is calculated. If a method is more reliable, then the correlation of the scores produced by the two halves will be high.
Fig.~\ref{fig:Reliability} shows the Spearman rank correlation ($\rho$) for half-sets obtained from rating scale and best--worst scaling data as a function of the available annotations in each half-set. 
It shows for each annotation set the split-half reliability using the full set of annotations (\mltpl{10}{N} per half-set) as well as partial sets obtained by choosing $k_{rs}$ annotations per term for rating scale (where $k_{rs}$ ranges from 1 to 10) or $k_{bws}$ annotations per 4-tuple for BWS (where $k_{bws}$ ranges from 1 to 5). 
The graph also shows BWS results obtained using \mltpl{1}{N}, \mltpl{1.5}{N}, and \mltpl{2}{N} unique 4-tuples.
In each case, the x-coordinate represents the total number of annotations in each half-set. 
Recall that the total number of annotations for rating scale equals \mltpl{k_{rs}}{N}, and for BWS it equals \mltpl{k_{bws}}{m}, where $m$ is the number of 4-tuples. Thus, for the case where $m = $\mltpl{2}{N}, the two methods are compared at points where $k_{rs} = $\mltpl{2}{k_{bws}}.

There are two important observations we can make from Fig.~\ref{fig:Reliability}.  
First, we can conclude that the reliability of the BWS annotations is very similar on the sets of \mltpl{1}{N}, \mltpl{1.5}{N}, and \mltpl{2}{N} annotated 4-tuples as long as the total number of annotations is the same. 
This means that in practice, in order to improve annotation reliability, one can increase either the number of unique 4-tuples to annotate or the number of independent annotations for each 4-tuple. 
Second,	annotations produced with BWS are more reliable than annotations obtained with rating scales. 
The difference in reliability is especially large when only a small number of annotations ($\leq5N$) are available. 
For the full set of more than 64K annotations (\mltpl{10}{N} = $\sim$32K in each half-set) available for both methods, the average split-half reliability for BWS is $\rho = 0.98$ and for the rating scale method the reliability is $\rho = 0.95$ (the difference is statistically significant, $p < .001$). 
One can obtain a reliability of $\rho = 0.95$ with BWS using just \mltpl{3}{N} ($\sim$10K) annotations in a half-set (30\% of what is needed for rating scale).\footnote{Similar trends are observed 
with Pearson's coefficient 
though the gap between BWS and RS results is smaller.}

\begin{table}[t]
\begin{center}
\small
\resizebox{0.49\textwidth}{!}{
\begin{tabular}{lrrr}
\hline \bf Term set & \bf \# terms & \multicolumn{1}{c}{\bf BWS} & \multicolumn{1}{c}{\bf RS} \\ \hline
all terms & 3,207 & .98 & .95\\[2pt]
single words & 1621 & .98 & .96\\[2pt]
all phrases & 1586 & .98 & .94\\
\ \ negated phrases & 444 & .91 & .78\\
\ \ \ \ pos.\@ phrases that have a negator & 83 & .79 & .17\\
\ \ \ \ neg.\@ phrases that have a negator & 326 & .81 & .49\\
\ \ modal phrases & 418 & .96 & .80\\
\ \ \ \ pos.\@ phrases that have a modal & 272 & .89 & .53\\
\ \ \ \ neg.\@ phrases that have a modal & 95 & .83 & .63\\
\ \ adverb phrases & 724 & .97 & .92\\
\hline
\end{tabular}
}
\end{center}
\vspace*{-3mm}
\caption{\label{split-half-modifiers} Average SHR 
for BWS and rating scale (RS) on different subsets of terms.}
\vspace*{-3mm}
\end{table}

Table~\ref{split-half-modifiers} shows the split-half reliability (SHR) on different subsets of terms. Observe that positive phrases that include a negator (the class that diverged most across BWS and rating scale), is also the class that has an extremely low SHR when annotated by rating scale. The drop in SHR for the same class when annotated with BWS is much less. Similar pattern is observed for other phrase classes as well, although to a lesser extent.
All of the results shown in this section, indicate that BWS surpasses rating scales on the ability to reliably rank items by sentiment, especially for phrasal items that are linguistically more complex.

\section{Conclusions}
We presented an experiment that directly compared 
the rating scale method of annotation with best--worst scaling.
We showed that, controlling for the total number of annotations, BWS produced significantly more reliable results.
The difference in reliability was more marked when about 5$N$ (or less) total annotations for an $N$-item set were obtained. 
BWS was also more reliable when used to annotate linguistically complex items such as phrases with 
negations and modals. 
We hope that these findings will encourage the use of BWS more widely to obtain high-quality annotations.

\section*{Acknowledgments}
We thank Eric Joanis and Tara Small for discussions 
on best--worst scaling and rating scales.

\bibliography{maxdiff}
\bibliographystyle{acl_natbib}

\appendix

\end{document}